\pdfoutput=1
\documentclass[11pt]{article}
\usepackage[]{acl} 
\usepackage{times}
\usepackage{latexsym}
\usepackage[T1]{fontenc}
\usepackage[utf8]{inputenc}
\usepackage{microtype}
\usepackage{inconsolata}

\usepackage{geometry} 
\usepackage{color}
\usepackage{amsfonts}
\usepackage{enumitem}
\usepackage{algorithm}
\usepackage{algpseudocode}
\usepackage{amsmath}
\usepackage{graphicx}
\usepackage{hyperref}
\usepackage{tikz,lipsum,lmodern}
\usepackage[most]{tcolorbox}
\usepackage{xcolor}
\usepackage{booktabs} 
\usepackage{array}    

\definecolor{codegreen}{rgb}{0,0.6,0}
\definecolor{codegray}{rgb}{0.5,0.5,0.5}
\definecolor{codepurple}{rgb}{0.58,0,0.82}
\definecolor{backcolour}{rgb}{0.95,0.95,0.92}

\usepackage{listings}
\usepackage{color}  

\lstdefinestyle{datastyle}{
    basicstyle=\footnotesize\ttfamily,
    backgroundcolor=\color{gray!5},  
    frame=single,  
    framerule=0pt,  
    framesep=5pt,  
    rulecolor=\color{black},  
    tabsize=2,
    breaklines=true
}

\author{
\bfseries Xiangpeng Wan\textsuperscript{1}, Haicheng Deng, Kai Zou\textsuperscript{1}, Shiqi Xu\textsuperscript{2}\\
\textsuperscript{1}ProtagoLabs \\
\textsuperscript{2}NetMind.ai \\
\{xiangpeng.wan, kz\}@protagolabs.com \\
denghaicheng777@gmail.com \\
michael@netmind.ai
}

\title{\vspace{-1cm}Enhancing the Efficiency and Accuracy of Underlying Asset Reviews in Structured Finance: The Application of Multi-agent Framework\vspace{-0.5cm}}

\date{May 2024}

\begin{document}
\maketitle

\begin{abstract}
\boldmath
    Structured finance, which involves restructuring diverse assets into securities like MBS, ABS, and CDOs, enhances capital market efficiency but presents significant due diligence challenges. This study explores the integration of artificial intelligence (AI) with traditional asset review processes to improve efficiency and accuracy in structured finance. Using both open-sourced and close-sourced large language models (LLMs), we demonstrate that AI can automate the verification of information between loan applications and bank statements effectively. While close-sourced models such as GPT-4 show superior performance, open-sourced models like LLAMA3 offer a cost-effective alternative. Dual-agent systems further increase accuracy, though this comes with higher operational costs. This research highlights AI’s potential to minimize manual errors and streamline due diligence, suggesting a broader application of AI in financial document analysis and risk management. The code is publicly accessible at \href{https://github.com/elricwan/Audit}{Audit}.
\end{abstract}

\section{Introduction}
Structured finance plays a pivotal role in financial engineering, involving the aggregation of diverse assets from typically large corporate borrowers into a single asset pool\cite{coval2009economics}. This pool is strategically restructured into tradable securities with varying priorities, including common instruments like Mortgage-Backed Securities (MBS), Asset-Backed Securities (ABS), and Collateralized Debt Obligations (CDOs). By reconfiguring the risk and return profiles of assets, these instruments provide borrowers with more flexible financing options with competitive rates while provide investors diverse investing options, thereby enhancing the efficiency and liquidity of capital markets.

However, the inherent complexity of these transaction structures, combined with the significant challenges in risk management, necessitates robust due diligence processes. Structured finance due diligence involves meticulous third-party evaluation of financial products and transactions to ensure their quality, compliance, and transparency—key factors in maintaining market confidence in these sophisticated financial instruments.

A crucial aspect of structured finance due diligence is the underlying asset review, which scrutinizes the assets comprising the structured finance pool\cite{arnholz2011offerings}. This review process is vital for verifying the completeness and accuracy of essential documents such as loan agreements, credit reports, financial information, property appraisal reports, and insurance policies. It also involves comprehensive compliance checks to ensure adherence to relevant legal and regulatory standards, and an assessment of asset quality through performance metrics such as default rates and repayment records. This meticulous process is traditionally conducted by professional audit teams or third-party service providers.

Despite the critical nature of these reviews, the process is plagued by challenges such as the voluminous nature of the documents, which can lead to significant time and resource consumption and human errors. Problems with document quality and consistency further complicate the accuracy and completeness of these reviews. Moreover, the presence of unstructured information within these documents often conceals essential insights and correlations, adding to the complexity of the task.

In response to these challenges, this paper proposes the integration of artificial intelligence (AI) technologies with human review processes to enhance the efficiency and accuracy of the review of underlying asset documents. AI has demonstrated considerable potential in improving process efficiency and accuracy in other knowledge-intensive fields, such as healthcare and law, through sophisticated data analysis and automated processing tasks. Yet, its application in structured finance due diligence remains relatively unexplored. This paper aims to bridge this research gap by using Auto asset-backed securities (auto ABS) as a case study to investigate the potential applications of AI in this context.

The contributions of this research include:
\begin{itemize}
    \item Systematic analysis of the challenges associated with reviewing underlying assets and the potential applications of AI, proposing solutions based on a multi-agent framework to increase automation, improve error detection and correction, and enhance the analysis of unstructured information.
    \item Empirical comparison of different large language models (LLMs) to evaluate their cost, efficiency, and error rate, demonstrating the benefits of AI agent methods.
    \item Development of an evaluation metrics system to assess the practical effects of AI technology in structured finance auditing.
\end{itemize}

\section{Related Works}

Structured finance\cite{culp2011structured} significantly contributes to the financial market by offering diversified investment options and flexible, competitive financing methods for borrowers, thereby enhancing market efficiency and liquidity\cite{jobst2007primer}. Fabozzi, Davis, and Choudhry (2006) highlight that structured financing allows borrowers to transfer asset risks to investors and capital markets, facilitating improved risk diversification and management\cite{fabozzi2006introduction}.

In structured finance, the review of underlying assets plays a crucial role across several dimensions in ensuring the quality, compliance, and transparency of transactions. It is foundational for risk management by allowing a detailed examination of underlying asset documents to identify and assess potential investment risks\cite{altomonte2014asset}. This process increases transparency, offering detailed information about the assets which aids investors in making informed decisions and understanding the risks involved in their investments\cite{neilson2022asset}. Such thorough reviews also promote trust by confirming the quality and compliance of assets, thereby enhancing investor confidence and market trust in ABS products\cite{pedio2019asset}. Additionally, these reviews are indispensable for ensuring that transaction structures comply with all applicable regulatory frameworks, helping to avoid potential legal and compliance risks and reinforcing the importance of regulatory compliance in structured finance transactions\cite{lamson2011developments}.

The application of multi-agent frameworks has been explored extensively across various domains, demonstrating their efficacy in tasks such as collective problem-solving\cite{du2023improving, wang2023unleashing,hao2023chatllm, liu2023training,park2023generative} and autonomous agent deployment\cite{park2023generative,gur2023real,qian2023communicative,zhou2023recurrentgpt}. Despite their potential, the use of AI agents, particularly large language models (LLMs) in structured finance, particularly within audit consulting services, remains under-explored.

This paper introduces a multi-agent framework utilizing LLMs to enhance the efficiency and accuracy of reviewing underlying asset documents in structured finance. The capabilities of this framework extend to automated document processing, sophisticated error detection and correction, and the analysis of unstructured information. By leveraging intelligent review tools, this approach not only streamlines the review process but also provides deeper insights, facilitating improved risk management and regulatory compliance in structured finance transactions.

\section{Methods}
\label{sec:others}

In the context of auto ABS, the underlying assets are typically auto loans or leases made to consumers to finance the purchase or lease of vehicles\cite{schultz2012auto}. Our primary task is to cross-verify information between a user's loan application and their bank statement. Essential details such as names, addresses, and balances need to be accurately matched. The key objective is to employ AI agents to extract targeted information from both documents efficiently, thereby reducing the time required for manual cross-verification.

Given the scarcity of open-sourced datasets containing bank and loan statements, we have developed an AI agent capable of generating this target data from basic inputs. Our workflow consists of several steps:

\textbf{Data Generation:} Initially, we generate random personal information, which is then used to create simulated bank statements and loan applications. This personal information is stored as labels for future verification.\\
\textbf{Document Processing:} A document reader agent converts these documents from PDF format into text.\\
\textbf{Information Extraction:} Subsequently, an extraction agent is employed to identify and pull the required information from the text.\\
\textbf{Accuracy Assessment:} The final step involves evaluating the extraction accuracy by comparing the extracted information against the true information stored as labels.

\begin{figure*}[t!]
	\vspace{0.2cm}
	\centerline{\includegraphics[width=0.9\textwidth]{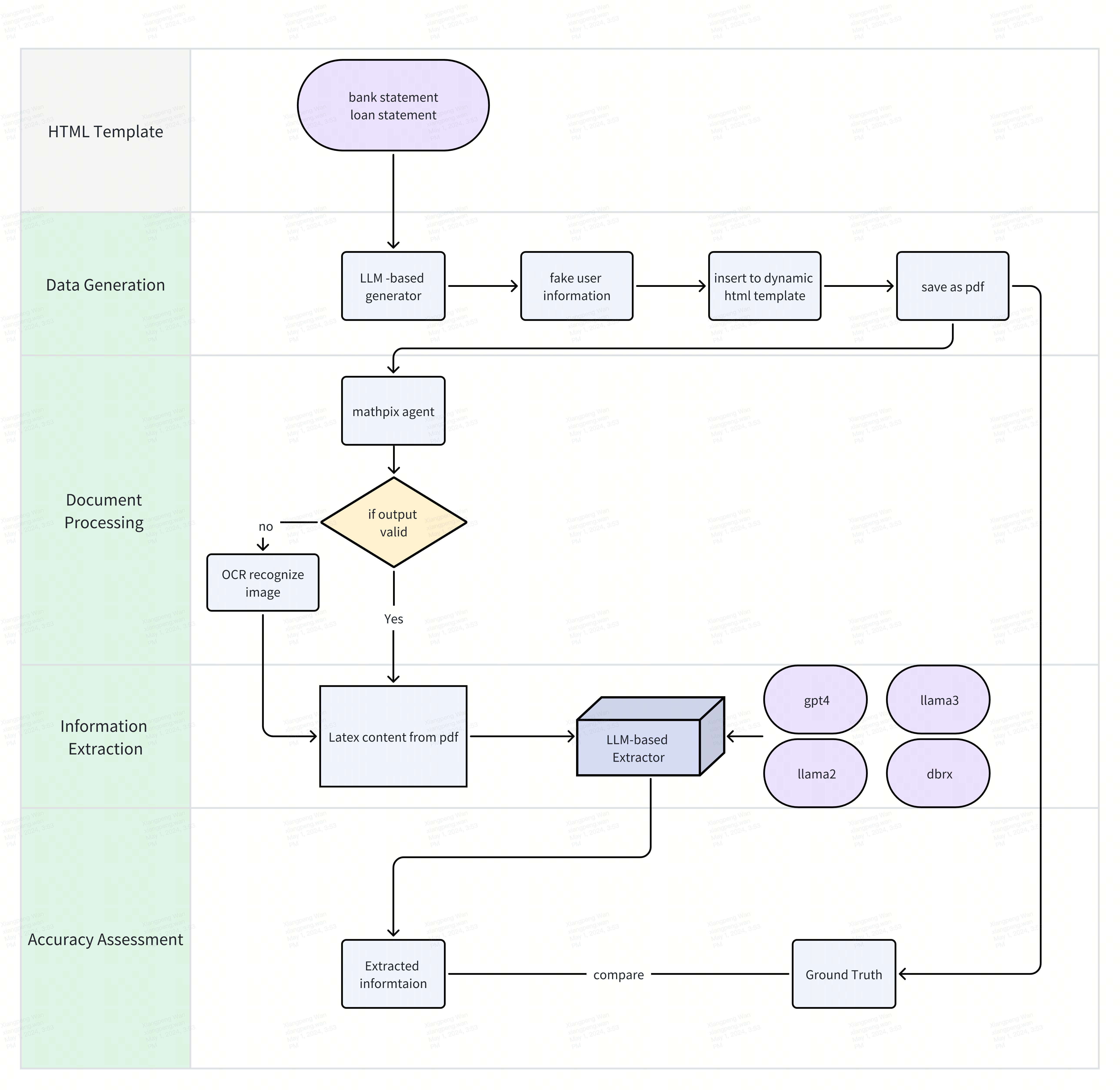}}\vspace{-0.1cm}
	\caption{\, The framework in competing the auto loan ABS task.   \normalsize}\label{Fig1}
	\vspace{-0.3cm}
\end{figure*}

The framework of this process is illustrated in Fig~\ref{Fig1}.

\subsection{Data}

In our experiments, we utilized three different bank statement templates and one loan statement template. We have made our HTML templates, along with the accompanying code, publicly accessible through our GitHub repository, available at \href{https://github.com/elricwan/Audit}{Audit}. This repository includes not only the code but also detailed usage examples within a demo, which are designed to assist users in effectively understanding and implementing our experimental procedures.

For our experiments, we employed an LLM-based agent to generate fictitious user information necessary for completing the bank statements and loan applications. It is important to note that some pieces of information, such as a person’s name, address, and statement date, are shared between the individual’s bank and loan statements. These details are enumerated in Listing~\ref{lst:fake_data}, and the generation prompts used for generating this information are specified in Listing~\ref{lst:gprompt}. The fake information we create, including names, addresses, statement dates, and transactions, is modeled after real audit scenarios where such data are crucial for cross-verifying with loan applications. In total, we generated 49 distinct bank statements and 49 corresponding loan statements.

\subsection{Extraction and Evaluation}

Our extraction agent employs LLM-based technologies. Currently, there are both open-sourced LLMs, like the LLAMA series \cite{touvron2023llama,touvron2023llama2} and Mistral \cite{jiang2023mistral}, and close-sourced models such as GPT-4 \cite{achiam2023gpt} and Gemini \cite{team2023gemini}. Close-sourced models often exhibit superior performance; however, open-sourced models provide advantages in cost and data privacy. In our experiments, we engaged with leading open-sourced models including LLAMA2, LLAMA3, and DBRX, comparing their performance against the close-sourced model GPT-4. Details of the extraction prompts are provided in Listing~\ref{lst:eprompt}.

After extracting data from bank statement and loan statement PDFs using various LLM-based agents, we evaluated the accuracy of the extracted data against ground truth values. To ensure precise evaluation results, we developed an evaluation agent based on GPT-4, which showed the highest performance among the models tested. The evaluation prompts are detailed in Listing~\ref{lst:evalprompt}. The performance of various models on the extraction task is systematically compared and detailed in Table~\ref{result}. This table provides a clear breakdown of how each model, both open-sourced and close-sourced, fares in terms of accuracy in extracting data from bank and loan statement PDFs.

\begin{table*}[ht]
\centering
\caption{Model Comparison on Extracting Accuracy}
\label{result}
\begin{tabular}{@{}lcccc@{}}
\toprule
\textbf{Metric}              & \textbf{gpt4} & \textbf{dbrx} & \textbf{llama3\_70b} & \textbf{llama2\_70b} \\ \midrule
\textbf{Bank Statement Overall Accuracy} & 0.99          & 0.91          & 0.93                & 0.92                \\
\textbf{Opening Balance}     & 0.98          & 0.94          & 1.00                & 1.00                \\
\textbf{Closing Balance}     & 0.98          & 0.96          & 1.00                & 0.94                \\
\textbf{Name}                & 1.00          & 1.00          & 1.00                & 1.00                \\
\textbf{Period Covered}      & 1.00          & 1.00          & 1.00                & 1.00                \\
\textbf{Address (Bank)}      & 0.97          & 0.67          & 0.67                & 0.67                \\
\textbf{Loan Statement Overall Accuracy} & 0.99          & 0.99          & 0.99                & 0.99                \\
\textbf{Address (Loan)}      & 1.00          & 1.00          & 1.00                & 1.00                \\
\textbf{Loan Amount}         & 0.98          & 0.98          & 0.98                & 1.00                \\
\textbf{Name (Loan)}         & 1.00          & 1.00          & 1.00                & 0.98                \\
\bottomrule
\end{tabular}
\end{table*}

Our evaluation results clearly demonstrate the superior performance of the close-sourced model, which achieved an overall accuracy of 0.99 across both bank and loan statement extractions. Similarly, the open-sourced model, such as LLAMA3, also displayed strong performance, with an accuracy of 0.93 for bank statements and 0.99 for loan statements. The primary challenge encountered with both models was extracting multi-line addresses from bank statements, where addresses typically span separate lines for the street name, city, and postcode.

It is important to note that these results were achieved using a single extraction agent in a zero-shot setting. However, when we deployed two agents to cross-verify information within the same document, we achieved a perfect accuracy rate of 100\% for both close-sourced and open-sourced models in our experiments. While this dual-agent approach increases both efficiency and cost compared to manual review, it is still significantly more economical and faster than employing individuals for the same task, who are more prone to errors.

Based on our findings, we can confidently assert that current multi-agent AI technologies are not just potentially but actually capable of significantly aiding individuals and companies in structured finance auditing.

\subsection{Cost analysis}

\begin{table*}[ht]
\centering
\caption{Pricing Per Model for Token Usage}
\label{price}
\begin{tabular}{@{}>{\raggedright\arraybackslash}p{2cm}>{\raggedright\arraybackslash}p{2.5cm}>{\raggedright\arraybackslash}p{3.5cm}>{\raggedright\arraybackslash}p{3.5cm}@{}}
\toprule
\textbf{Model} & \textbf{Provider} & \textbf{Price per 1M Tokens (Input)} & \textbf{Price per 1M Tokens (Output)} \\ \midrule
gpt4           & Open-AI           & \$10.00                             & \$30.00                               \\
llama3\_70b    & Replicate         & \$0.65                              & \$2.75                                \\
llama2\_70b    & Replicate         & \$0.65                              & \$2.75                                \\
dbrx     & Together.ai       & \$1.20                              & \$1.20                                \\
\bottomrule
\end{tabular}
\end{table*}

In Table~\ref{price}, we outline the pricing for using various models. Typically, a monthly bank statement contains transactions, balances, fees, and occasionally promotional material, which generally totals between 500 and 2,000 tokens. On the other hand, a loan statement, usually shorter, ranges from about 300 to 1,500 tokens. Including prompts, the total input length for processing rarely exceeds 5,000 tokens, while the output, containing all key information, generally comprises fewer than 100 tokens.

Based on these estimates, the cost of processing a document with the gpt4 model is approximately \$0.05 per document. For the llama models, the cost is around \$0.00325 per document, and for the dbrx model, it is approximately \$0.006 per document.

\section{Conclusion}
This paper examined the application of multi-agent AI frameworks to improve the review processes for underlying asset documents in structured finance. Our findings demonstrate that both open-sourced and close-sourced LLM-based agents can effectively automate data extraction tasks, achieving high levels of accuracy in cross-verifying information between loan applications and bank statements.

While close-sourced models showed slightly better performance, the open-sourced models also performed robustly and at a lower cost. The use of dual-agent systems further enhanced accuracy, although at the expense of higher operational costs. Despite these costs, AI systems offer a reliable and cost-effective alternative to manual reviews, which are prone to error.

Future research should explore ways to enhance the scalability and cost-efficiency of these technologies, and to extend their application to broader financial document types. As AI continues to advance, its integration into structured finance auditing could significantly improve both efficiency and reliability in financial risk management and compliance.

\textbf{Limitation.} 
The current study may be limited to specific types of financial documents. The ability of the AI models to adapt to different formats or less structured data has not been extensively tested. Utilizing AI in financial auditing must comply with stringent regulatory and privacy standards. The adaptation of AI tools must ensure they do not violate privacy laws, which can vary widely across jurisdictions.

\textbf{Future.} 
Future research could focus on testing and adapting the AI models to a wider variety of financial documents, including those with less structured formats. This would help assess the robustness and versatility of the AI systems across different financial contexts.

\bibliography{structured_finance/acl_latex}

\section*{Appendix: Prompt}

The following prompt was used during the evaluation of large language models in our experiments. This prompt defines different roles of our agents, detailing the task requirements and expected output format for solving completing the task.

\begin{lstlisting}[caption={Bank and Loan Data},label=lst:fake_data]
bank_data = {
    "Account_Number": "123-456-789",
    "Statement_Date": "2024-03-01",
    "Period_Covered": "2024-02-01 to 2024-02-29",
    "name": "John Doe",
    "address_line1": "2450 Courage St, STE 108",
    "address_line2": "Brownsville, TX 78521",
    "Opening_Balance": "175,800.00",
    "Total_Credit_Amount": "510,000.00",
    "Total_Debit_Amount": "94,000.00",
    "Closing_Balance": "591,800.00",
    "Account_Type": "Savings",
    "Number_Transactions": "10",
    "transactions": [
        {"Date": "2024-03-01", "Description": "Coffee Shop", "Credit": "$50.00", "Debit": "-$5.00", "Balance": "$995.00"},
        {"Date": "2024-03-01", "Description": "Online Purchase", "Credit": "$121.51", "Debit": "-", "Balance": "$1,116.51"}, 
        {"Date": "2024-03-02", "Description": "Coffee Shop", "Credit": "$143.06", "Debit": "-", "Balance": "$1,259.57"}, 
        {"Date": "2024-03-03", "Description": "Utility Bill", "Credit": "-", "Debit": "-$60.72", "Balance": "$1,198.85"}, 
    ]
}

loan_data = {
    "title": "Loan Application Form",
    "form_title": "Please Fill Out the Loan Application",
    "form_action": "/submit-application",
    "applicant": {
        "first_name": "Jane",
        "last_name": "Doe",
        "ssn": "987-65-4321",
        "dob": "1990-05-15",
        "email": "jane.doe@example.com",
        "phone": "555-6789",
        "address": "123 Elm Street, Yourtown, YS",
        "marital_status": "Single",
        "employment_status": "Employed",
        "employer_name": "YourCompany",
        "annual_income": 50000,
        "other_income": 5000,
        "monthly_expenses": 2000
    },
    "marital_statuses": ["Single", "Married", "Divorced", "Widowed"],
    "employment_statuses": ["Employed", "Unemployed", "Self-Employed", "Retired"],
    "loan_details": {
        "amount": 25000,
        "purpose": "Home Renovation",
        "term": 10,
        "interest_rate": "5.5%"
    },
    "loan_purposes": {
        "Home Purchase": "Home Purchase",
        "Home Renovation": "Home Renovation",
        "Debt Consolidation": "Debt Consolidation",
        "Education": "Education",
        "Other": "Other"
    }
}

\end{lstlisting}

\begin{lstlisting}[caption={Generation Prompt},label=lst:gprompt]
BANK_INFO = [
    {"role" : "system",
     "content": """Now, you are a Banking assistant who can help user to generate logical user information for bank statement.
Here is a sample of information that you need to follow:
{{
    "Account_Number": "123-456-789",
    "Statement_Date": "2024-03-01",
    "Period_Covered": "2024-02-01 to 2024-02-29",
    "name": "John Doe",
    "address_line1": "2450 Courage St, STE 108",
    "address_line2": "Brownsville, TX 78521",
    "Opening_Balance": "175,800.00",
    "Total_Credit_Amount": "510,000.00",
    "Total_Debit_Amount": "94,000.00",
    "Closing_Balance": "591,800.00",
    "Account_Type": "Savings",
    "Number_Transactions": "10",
    "transactions": [
        {{"Date": "2024-03-01", "Description": "Coffee Shop", "Credit": "$50.00", "Debit": "-$5.00", "Balance": "$995.00"}},
        {{"Date": "2024-03-01", "Description": "Online Purchase", "Credit": "$121.51", "Debit": "-", "Balance": "$1,116.51"}}, 
        {{"Date": "2024-03-02", "Description": "Coffee Shop", "Credit": "$143.06", "Debit": "-", "Balance": "$1,259.57"}}, 
        {{"Date": "2024-03-03", "Description": "Utility Bill", "Credit": "-", "Debit": "-$60.72", "Balance": "$1,198.85"}}, 
    ]
}}
## You must follow all the requirements to modify the draft:
    1. You must generate information given in the sample, including "Account_Number", "Statement_Date", etc.  
    2. You must generate several "transactions", the number could vary.
    3. You must generate logical values, the "Statement_Date", "Period_Covered" and "Date" in "transactions" must be resaonable.
    4. You must generate unique user information, not seen in the history. 

## About the output:
    Your output must be a json file containing a python dictionary to store the extracted information in the format looks like the sample above. 
    You must follow all requirements listed above. 
    Your output must contain the json file quoted by "```json" and "```"

    """},
    {"role": "user",
    "content": """
        Here is the history: {history}. 
"""}]

LOAN_INFO = [
    {"role" : "system",
     "content": """Now, you are a loan application assistant who can help user to generate logical user information for loan application. 
    
     Here is a sample data that you need to follow:
    {{
        "title": "Loan Application Form",
        "form_title": "Please Fill Out the Loan Application",
        "form_action": "/submit-application",
        "applicant": {{
            "first_name": "Jane",
            "last_name": "Doe",
            "ssn": "987-65-4321",
            "dob": "1990-05-15",
            "email": "jane.doe.fake@example.com",
            "phone": "555-6789",
            "address": "123 Elm Street, Yourtown, YS",
            "marital_status": "Single",
            "employment_status": "Employed",
            "employer_name": "YourCompany",
            "annual_income": 50000,
            "other_income": 5000,
            "monthly_expenses": 2000
        }},
        "employment_statuses": ["Employed", "Unemployed", "Self-Employed", "Retired"],
        "loan_details": {{
            "amount": 25000,
            "purpose": "Home Renovation",
            "term": 10,
            "interest_rate": "5.5%"
        }},
        "loan_purposes": {{
            "Home Purchase": "Home Purchase",
            "Home Renovation": "Home Renovation",
            "Debt Consolidation": "Debt Consolidation",
            "Education": "Education",
            "Other": "Other"
        }}
    }}
## You must follow all the requirements to modify the draft:
    1. You must generate the same structure dictionary as the sample, including all the keys and values.
    2. You must generate complete dictionary, each key should have a corresponding value.
    3. You would be given only part of the user information, you must use the information to fill the generated dictionary.
    4. You must generate logical values for those information not given in the user information.
    
## About the output:
    Your output must be a json file containing a python dictionary to store the extracted information in the format looks like the sample above. 
    You must follow all requirements listed above. 
    Your output must contain the json file quoted by "```json" and "```"

    """},
    {"role": "user",
    "content": """
        Here is the user information: {user_information}. 
"""}]
\end{lstlisting}

\begin{lstlisting}[caption={Extract Prompt},label=lst:eprompt]
AUDIT = [
    {"role" : "system",
     "content": """Now, you are a Audit assistant who can help user to extract information from text.
    ## You must follow all the requirements to modify the draft:
        1. You must extract the name of person from the text, including first and last name.
        2. You must extract the period_covered from the text, if given.
        3. You must extract the address from the text, if given.
        4. You must extract the Opening Balance from the text, if given.
        5. You must extract the Closing Balance from the text only if given.
        6. You must extract the loan amount from the text only if the text is about loan application.
    
    ## About the output:
        Your output must be a json file containing a python dictionary to store the extracted information in the format looks like this: 
        
        {{
            "name": "xxx",
            "period_covered": "xxx",
            "address": "xxx",
            "period_covered": "xxx",
            "opening_balance": "xxx",
            "closing_balance": "xxx",
            "loan_amount": "xxx",
        }}
        You must follow all requirements listed above. 
        Your output must contain the json file quoted by "```json" and "```"

    """},
    {"role": "user",
    "content": """
        The given text is:

    {text}
"""}]

\end{lstlisting}

\begin{lstlisting}[caption={Evaluation Prompt},label=lst:evalprompt]
EVALUATION = [
    {"role" : "system",
     "content": """Now, you are a evaulator who can help user to determine the accurate rate of prediction file compared to the true file.
    Here is a sample of prediction file:
    {{
        "name": "xxx",
        "period_covered": "xxx",
        "address": "xxx",
        "period_covered": "xxx",
        "opening_balance": "xxx",
        "closing_balance": "xxx",
        "loan_amount": "xxx",
    }}
## You must follow all the requirements to complete the task:
    1. You must compare each items that exists with valid value on prediction file to the true file. 
    2. If there are both first name and last name in true file, you need to combine them together as name. If there are several address in true file, you need to combine them together as address.
    3. If the item with valid value exists in the prediction file but not in the true file, it counts as incorrectly predicted.
    4. You must record the items that are correctly predicted.
    5. You must record the items that are incorrectly predicted.
    6. You must count the number of items that are correctly predicted.
    7. You must count the number of items that are incorrectly predicted.
    8. You must calculate the accuracy of the prediction. 

## About the output:
    Your output must be a json file containing a python dictionary to store the result, the format follows this:
    {{
        "correctly_predicted_items": ["xxx", "xxx", "xxx"],
        "incorrectly_predicted_items": ["xxx", "xxx", "xxx"],
        "correctly_predicted": "xxx",
        "incorrectly_predicted": "xxx",
        "accuracy": "xxx",
    }}
    You must follow all requirements listed above. 
    Your output must contain the json file quoted by "```json" and "```"

    """},
    {"role": "user",
    "content": """
        Here is the prediction file: {prediction}. 
        Here is the true file: {true}.
"""}]
\end{lstlisting}

\end{document}